\documentclass[conf]{new-aiaa}
\usepackage[utf8]{inputenc}

\usepackage{graphicx}
\usepackage{amsmath}
\usepackage[version=4]{mhchem}
\usepackage{siunitx}
\usepackage{longtable,tabularx}
\newcommand{\comment}[1]{}
\usepackage{subfig}
\usepackage{placeins}

\usepackage{float}
\usepackage{algorithm}
\usepackage{algpseudocode}

\title{Differentiable Rendering for Pose Estimation in Proximity Operations}

\author{Ramchander Rao Bhaskara\footnote{Graduate Student, Aerospace Engineering, Texas A\&M University, College Station, TX 77843.}}
\affil{Texas A\&M University, College Station, TX 77843}
\author{Roshan Thomas Eapen\footnote{Assistant Professor, Aerospace Engineering, 
Pennsylvania State University, State College, PA 16801.}}
\affil{Pennsylvania State University, State College, PA 16801}
\author{Manoranjan Majji\footnote{Associate Professor, Aerospace Engineering, Texas A\&M University, College Station, TX 77843.}}
\affil{Texas A\&M University, College Station, TX 77843}

\begin{document}

\maketitle

\begin{abstract}
Differentiable rendering aims to compute the derivative of the image rendering function with respect to the rendering parameters. This paper presents a novel algorithm for 6-DoF pose estimation through gradient-based optimization using a differentiable rendering pipeline. We emphasize two key contributions: (1) instead of solving the conventional 2D to 3D correspondence problem and computing reprojection errors, images (rendered using the 3D model) are compared only in the 2D feature space via sparse 2D feature correspondences. (2) Instead of an analytical image formation model, we compute an approximate local gradient of the rendering process through online learning. The learning data consists of image features extracted from multi-viewpoint renders at small perturbations in the pose neighborhood. The gradients are propagated through the rendering pipeline for the 6-DoF pose estimation using nonlinear least squares. This gradient-based optimization regresses directly upon the pose parameters by aligning the 3D model to reproduce a reference image shape. Using representative experiments, we demonstrate the application of our approach to pose estimation in proximity operations.  

\comment{
to converge to an observed image geometry using a few reference images synthesized at inference time. Our algorithm computes an approximate numerical gradient of the rendering process using image features gathered from multiviewpoint renders attained from small variations in the pose neighborhood. The gradients are propagated through the images for pose estimation using nonlinear least-squares based optimization posed to minimize the feature misalignment errors. Representative experiments are carried out utilizing the International Space Station (ISS) to evaluate the efficacy of the proposed online differentiable rendering framework in proximity operations.}
\end{abstract}

\section{Introduction}

\comment{This paper proposes a model-based approach for determining pose from a single image.}

Space-borne navigation relies on accurate estimation of relative position and attitude, known as relative pose, between a target and a chaser spacecraft. Advances in computer vision have propelled both the utility and ubiquity of vision-based relative navigation, using visual sensors to estimate relative pose \cite{du2009pose,cropp2000estimating,zhang2013vision}. Vision-based pose estimation is widely implemented in positioning applications such as autonomous rendezvous and docking \cite{6978896}, proximity operations \cite{junkins1999vision}, and terrain relative navigation \cite{verras2021vision, bhaskara2022fpga}. In cooperative proximity operations, the computation of pose using image feature coordinates is classically achieved by solving \textit{Perspective-n-Point} (PnP) problem \cite{fischler1981random}. The geometric configuration of the features of the target object is usually known, either from deployed fiducial markers or from transmission of state information \cite{opromolla2017review, sung2020interferometric}.

However, the estimation of the relative pose between an uncooperative target and a chaser is a more involved process. Without assistance from the target, the chaser must lead the efforts in feature acquisition and tracking for correspondence. Also, because 2D images lack depth information, pose cannot be determined using a single image. Monocular or stereo vision pipelines use at least two frames to first estimate the 3D locations of the feature points and then determine the relative pose \cite{eapen2022narpa}. However, the estimation of 3D feature coordinates is associated with its own errors and therefore could potentially corrupt the pose estimation accuracy. If the geometric model of a target is known in advance or built online \cite{wong2018photometric}, the pose can be determined by mapping the feature points to the 3D locations on the target model \cite{sharma2018robust}. Otherwise, in a model-free approach, sparse 3D point clouds of a target can be constructed using Structure from Motion (SfM) \cite{capuano2020monocular}.  

Traditionally, in the model-based relative navigation approach, pose is determined through the PnP algorithm after establishing correspondences between the 2D feature points and the 3D model points. This approach has two major drawbacks:
\begin{enumerate}
    \item \textbf{Robustness}. Geometric methods, PnP or POSIT \cite{dementhon1995model}, solve the estimation problem as a function of pose only. However, rapid changes in illumination conditions could adversely affect the solution of feature correspondence and pose estimation problems. 
    \item \textbf{Accuracy limitations}. Establishing 2D-3D correspondence as an initial step adds to the errors in pose estimation. Numerical approximations and pixel interpolations limit the precision of pose estimates.  
\end{enumerate}

\comment{
In this paper, we propose a differentiable rendering pipeline for model-based pose estimation. It has the following features:
\begin{enumerate}
    \item Determines the pose from a single image. 
    \item Online learning of an image formation model that is inherently sensitive to local illumination conditions. 
    \item Optimization in the 2D feature space without requiring to solve any correspondence problem.  
\end{enumerate}
}

Mitigating these drawbacks, we build on our previous work \cite{eapen2022narpa} and propose a differentiable rendering pipeline for pose estimation. 
\comment{The pipeline utilizes ray tracing for online learning of the image synthesis gradient.

For the testing and validation of computer vision pipelines, it is critical to simulate physically realistic navigation scenarios using computer graphics \cite{abderrahim2005experimental}.} 

\subsection{Differentiable Rendering}
Ray tracing tools aim to render physically realistic images of an object given its 3D geometry and texture under specified camera and illumination properties \cite{jakob2013mitsuba, eapen2022narpa}. This process is also termed \textit{forward rendering}. On the other hand, computer vision is viewed as an inverse graphics process to search for the parameters of a model that are used to render the image in the first place \cite{baumgart1974geometric,loper2014opendr}. In essence, inverse graphics is an optimization approach to infer the 3D geometry and scene properties from an image, so that a graphics engine can reconstruct the observed image. In addition, \textit{Differentiable Rendering} (DR) is a reverse engineering technique to model changes in the output of the rendered image with respect to changes in the parameters used to render the image \cite{kato2020differentiable, wu2020analytical,park2020latentfusion}. In short, differentiable rendering is a technique of analysis by synthesis to estimate image generation parameters using iterative optimization rules \cite{mansinghka2013approximate}. 

OpenDR framework, a first general-purpose differentiable rendering solution, is proposed by Loper et al. \cite{loper2014opendr}. It embeds a forward graphics model to render images from scene parameters and to generate the image derivatives with respect to the model parameters using approximate numerical methods. Our operation is coherent in principle with OpenDR wherein for a given model of the scene geometry and image features, we define an observation error function and iteratively minimize the image differences by differentiating it with respect to the model parameters. We distinguish ourselves from OpenDR-like frameworks and other pose estimation pipelines in at least one of the following ways:

\begin{enumerate}
    \item The image formation model supplies pixel measurements as a function of pose parameters (only) but is sensitive to local illumination conditions. 
    \item Online learning of the rendering function's Jacobian for data-driven optimization. Data generated by uniform sampling of pose parameters along a current pose axis.  
    \item Optimization in the 2D feature space without requiring to solve any 3D correspondence problem.
    \item Emphasis on physically accurate image rendering of virtual scenes to prioritize true-to-physics navigation scenarios. 
\end{enumerate}

Other state-of-the-art pose estimation frameworks are based on supervised learning methods \cite{kato2018neural,park2020latentfusion,eslami2018neural}. The neural rendering frameworks are a set of very robust pose estimation pipelines for terrestrial applications, but require extensive training under varied illumination conditions \cite{kehl2017ssd,xiang2017posecnn,rad2017bb8,sharma2018pose,park2019towards}. 
Offline learning frameworks for spacecraft navigation through neural nets are on the trickier end due to a) rapidly changing illumination conditions affected by occlusions due to fast relative motion between the chaser and the target satellite, and b) lack of extensive and elaborate training data sets for reliable pose estimation of unseen or partially observable objects. 

In this paper, we present the estimation of the 6-DoF pose from a single image by posing the forward rendering process as a function of the pose parameters. Our process does not assume constancy in illumination but relies on extraction of a set of common image features in each iteration towards convergence. The core of our method is an online learning approach to approximately compute the gradient of the rendering process with respect to the pose parameters. The online learning process is facilitated by systematic pose perturbations. The framework extracts feature points from RGB images with known poses and tracks the respective features in the images generated in the perturbed pose neighborhood. A local gradient (Jacobian) built from these feature differences iteratively refines the pose to fit the renderings of the 3D model to the reference image. 

The remainder of this paper is organized as follows. In Section \ref{sec:prelims}, a brief overview of the differentiable rendering pipeline is provided. Section \ref{sec:jacobian} describes the method for estimating the Jacobian of the rendering process, and Section \ref{sec:optim} presents the nonlinear optimization problem for 6-DoF pose estimation. The results for the validation of the proposed pipeline to estimate the pose in proximity operations are presented in Section \ref{sec:res}.

\comment{
the overview of our differentiable rendering pipeline for view reconstruction is described. We emphasize one-shot pose estimation by leveraging the reconstruction and rendering pipeline. Two, we present experiments and the results of pose estimation and view reconstruction using our novel framework. Our conclusion highlights the efficacy as well as gaps in our differentiable rendering process while presenting future plans of action.
}
\section{Preliminaries of Differentiable Rendering} \label{sec:prelims}

\begin{figure}[hbt!] 
    \centering
    \includegraphics[width=0.85\textwidth]{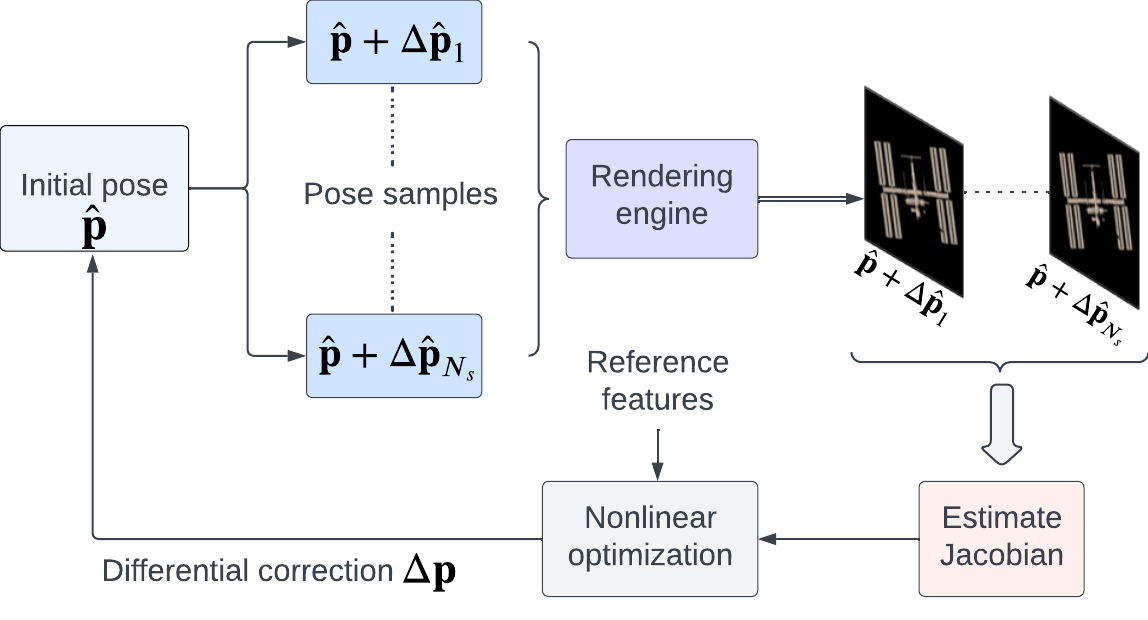} 
    \caption{Differentiable rendering pipeline for pose estimation: 1) the framework utilizes a rendering engine to synthesize an RGB image of the object at an initial pose. Arbitrary views of the object are generated via small perturbations in the pose neighborhood. An approximate gradient of the image is numerically computed using the registered poses and their respective image renderings. The gradient is utilized to optimize the view to reproduce the 2D view geometry and estimate 6-DoF relative pose.}
    \label{fig:dr_overview}
    \end{figure}

Given a single RGB reference image consisting of a target object, we present an end-to-end pipeline for novel 2D view synthesis and 6-DoF relative pose (rotation and translation) estimation. Using the object model and a valid initial guess for the pose, we render an image and extract a set of reference features that match with those of the reference image. The reference features are tracked in the neighborhood of the pose guess to locally approximate the image rendering gradient with respect to the pose parameters. This provides a direction for gradient-based optimization to minimize misalignment error for precise 2D view reconstruction and pose estimation. Our overall architecture is highlighted in Fig. \ref{fig:dr_overview}. The process of forward rendering and image formation kinematics are discussed below.

\subsection{Image Formation Model}

In this work, we utilize a perspective pinhole camera model to transform object geometry to view space. In perspective projection, given $\textit{n}$ 3D reference points $\mathbf{a}_i, \, \textit{i} = 1,2,...,n$, in the object reference frame, and their transformed 3D coordinates $\mathbf{b}_i$, in the view space, we seek to retrieve the proper orthogonal rotation matrix $\mathbf{R}$ and the translation vector $\mathbf{t}$, of the camera relative to a target satellite. The projection transformation between the 3D reference points and the corresponding view space 3D coordinates is given by 
\begin{equation} \label{projectionTransformation1}
    \mathbf{b}_i = \mathbf{R}\mathbf{a_i} + \mathbf{t},\hspace{0.5cm} i=1,2,...,n
\end{equation}
\noindent where $\mathbf{R}^T\mathbf{R} = I$ and $\text{det}(\mathbf{R}) = 1$.

The perspective projection model extends the transformation in Eq. (\ref{projectionTransformation1}) from 3D view space coordinates to their corresponding 2D homogeneous image projections $\mathbf{l}_i = [u_i \; v_i \; 1]^T$ for given intrinsic camera parameters as
\begin{equation} \label{proTrans2}
s \mathbf{l}_i = K \: [R \; | \; \mathbf{t}]\, [\mathbf{a}_i \quad 1]^T
\end{equation}

\noindent where $s$ denotes the depth factor for the $i$-th point and $K$ is the matrix of calibrated intrinsic camera parameters corresponding to axis skew $\gamma$, aspect ratio scaled focal lengths $(f_x$, $f_y)$, and principal point offset $(x_0, y_0)$:
\begin{equation} \label{Kmat}
K = \begin{bmatrix}
 {f_x} & \gamma & x_0 \\
 0 & {f_y} & y_0 \\
 0 & 0 & 1
\end{bmatrix}
\end{equation} 

The attitude kinematics used in this paper utilizes classical Rodrigues parameters (CRPs)  \cite{schaub1996stereographic} (Gibbs vector) to represent the orientation matrix $\mathbf{R}$. CRPs facilitate posing an optimization problem via polynomial system solving, free of any trigonometric functions. Let vector $\mathbf{q} = [q_1\; q_2\; q_3]^T$ denote the CRPs, the rotation matrix $\mathbf{R}$ in terms of $\mathbf{q}$ can be obtained using the Cayley transform as: 
\begin{equation} 
\mathbf{R} = (I + [\mathbf{q}\times])^{-1} (I - [\mathbf{q}\times])
\end{equation}
where $I \in {\rm I\!R}_{3 \times 3}$ is an identity matrix, and operator $[\mathbf{q}\times] \in {\rm I\!R}_{3 \times 3}$ converts a vector into a skew-symmetric matrix of the form: 
\begin{equation}
[\mathbf{q}\times] = \begin{bmatrix}
0 & -q_3 & q_2 \\
q_3 & 0 & -q_1 \\
-q_2 & q_1 & 0
\end{bmatrix}
\end{equation}  

\noindent and the resulting parameterization in vector form is 
\begin{equation} \label{CRPDCM}
    \mathbf{R} = \frac{1}{1+\mathbf{q}^T\mathbf{q}} \left( (1 - \mathbf{q}^T\mathbf{q}) [{I}_{3 \times 3}] + 2 \mathbf{q} \mathbf{q}^T - 2 [\mathbf{q} \times]\right)
\end{equation}

\comment{
\begin{equation} 
    \mathbf{R} = \frac{1}{1+\mathbf{q}^T\mathbf{q}} \begin{bmatrix}
    1 + q_1^2-q_2^2-q_3^2 & 2(q_1q_2 + q_3) & 2(q_1q_3 - q_2) \\
    2(q_2q_1 - q_3) & 1 - q_1^2+q_2^2-q_3^2 &  2(q_2q_3 + q_1) \\
     2(q_3q_1 + q_2) & 2(q_3q_2 - q_1) & 1 + q_1^2-q_2^2+q_3^2
    \end{bmatrix}
\end{equation}
}

In what follows, we present several tools to construct a nonlinear optimization approach for the estimation of pose parameters $\mathbf{p} = [\mathbf{q} \quad \mathbf{t}]^T$. We begin with a brief description of the rendering process and direct the readers to our previous work \cite{eapen2022narpa} for a detailed description of the rendering algorithm.

\subsection{Forward Rendering}
\begin{figure}[hbt!] 
    \centering
    \includegraphics[width=0.85\textwidth]{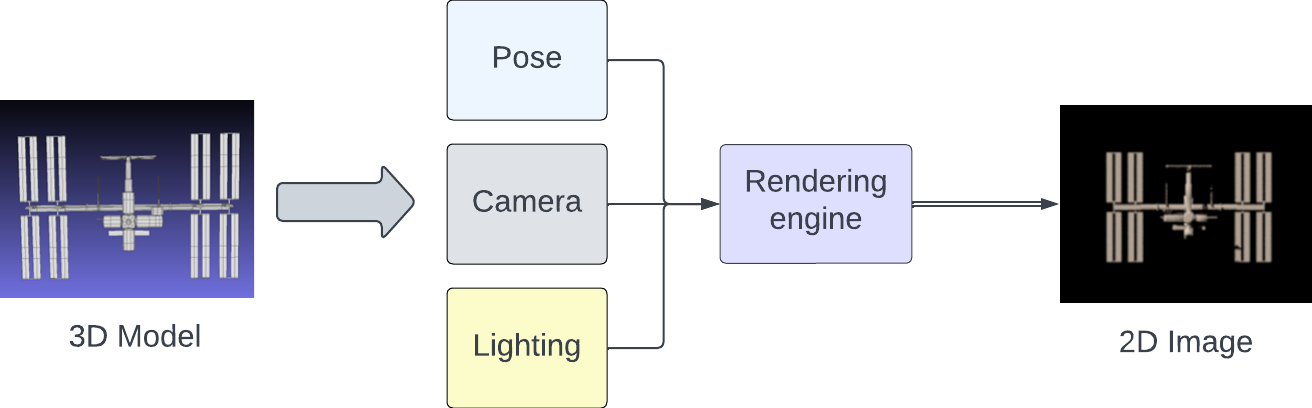} 
    \caption{Forward Rendering Process: With the 3D model of the object, relative pose of the camera, and the scene illumination properties as the inputs, the rendering engine synthesizes RGB images at the output.}
    \label{fig:fr_overview}
    \end{figure}

The forward rendering process utilizes a ray-tracing based rendering engine that takes the 3D object model as input and renders a 2D image as described in Fig. \ref{fig:fr_overview}. The virtual scene is constructed with the description of view geometry i.e., the relative pose between the camera and the object, the camera instrinsic parameters, and illumination attributes. We deploy physically based ray-tracing engine for rendering images that are physically accurate and empirically valid. Our pipeline is compatible for deployment with the Mitsuba \cite{jakob2013mitsuba} engine as well as our own engine called Navigation and Rendering Pipeline for Astronautics (NaRPA) \cite{eapen2022narpa}. In the next section, we present a method of estimating the gradient of the image rendering process with respect to the pose parameters.

\section{Estimating the Jacobian} \label{sec:jacobian}

The image formation model is a nonlinear geometric relationship construct $\mathbf{x}_t = \mathbf{f}(\mathbf{p}_t)$ between the degrees of freedom of the system $\mathbf{p}_t$ and its corresponding output $\mathbf{x}_t$ for any time $t$. Similar to the concept of inverse kinematics, the objective of differentiable rendering approach is to estimate $\mathbf{p}_t$ from $\mathbf{x}_t$. A global search to identify the optimal $\mathbf{p}$ might be intractable for real-time applications due to the high-dimensional search space and highly nonlinear $\mathbf{f}$. Alternately, the nonconvex and high-dimensional pose estimation problem is desirable to be optimized locally and serviceable using online numerical computations. 

\subsection{Method}
Consider the set of 6-DoF camera pose parameters $\mathbf{p}=(p_i)_{1 \leq i \leq 6}$ describe the image feature measurements $\mathbf{x} = (x_j)_{1 \leq j \leq m}$ through the image formation process $\mathbf{x}=\mathbf{f}(\mathbf{p})$. The goal is to estimate $\mathbf{p}$ from the observations $\mathbf{x}$ by solving a generalized inverse problem. 

The nonlinear and non-invertible process $\mathbf{f}$ is only explicitly modeled as a function of $\mathbf{p}$ but is implicitly sensitive to illumination as well as the intrinsics of the capturing camera. We bypass the analytical description of the image formation model because of the complexities conditioned on not just the relative pose, but also the modeling of lens and illumination sources. Therefore, we seek a local numerical solution $\Delta \mathbf{p} = \mathbf{M} \Delta \mathbf{x}$ for some $\Delta \mathbf{x}$ in the neighborhood of zero and $\mathbf{M}$ to be computed. In this linear approximation, our approach allows to take the implicit parameters into account, in a small linear neighborhood, without supplementary computations. Ignoring higher-order terms, from the Taylor series expansion of $\mathbf{x}$,
\begin{equation} \label{eq:jac_linear}
    \Delta \mathbf{x} = \mathbf{J}_f \Delta \mathbf{p}
\end{equation}

\noindent $\mathbf{J}_f$ being the Jacobian matrix that acts as a local gradient of the image formation model, in the neighborhood of some initial pose $\mathbf{p}$. 

In this section, we propose a learning approach to numerically compute $\mathbf{J}_f$. This is achieved by generating a random sample of $N_s$ increments $\Delta \mathbf{p}$, producing a $n \times N_s$ matrix of perturbations $\mathbf{p}+\Delta \mathbf{p}$. The resulting $m \times N_s$ matrix of corresponding measurement variations is calculated through $\mathbf{f}$ as $\Delta \mathbf{x} = \mathbf{f}(\mathbf{p}+ \Delta\mathbf{p}) - \mathbf{f}(\mathbf{p})$. Then, the Jacobian of the image formation process is evaluated by resolving the linear system in Eq. (\ref{eq:jac_linear}) in the least-squares sense. If the $n \times N_s$ matrix composed of pose samples has a rank equal to $n$, the Jacobian $\mathbf{J}_f $ is computed from the right inverse of this matrix. This process is described in the next section. 

\subsection{Realization}
In image based visual servoing, Jacobian is often used to design control laws for moving an end-effector to a desired pose based on visual feedback. Learning techniques for numerical estimation of the Jacobian are popularized to mitigate the need for a priori knowledge of kinematic structure and calibration parameters \cite{hosoda1994versatile}. Traditionally, online learning to estimate the Jacobian based on visual features is aimed at tracking a target pose and not necessarily estimating the pose \cite{jagersand1997experimental}. Learning the inverse Jacobian has also been shown to result in much better performance than inverting an estimate of the Jacobian \cite{lapreste2004efficient}. From all of these previous studies, if a pose estimate is to be drawn, a finite difference scheme may be deployed to correct the pose parameters. This work uses the estimated Jacobian to compute a maximum likelihood estimate for the pose parameters that iteratively minimizes the feature errors. In any case, the learning model relies on the scheme of the pose perturbations for the Jacobian approximation. 

\subsubsection{Pose Sampling}

\begin{figure}[hbt!] 
    \centering
    \includegraphics[width=0.5\textwidth]{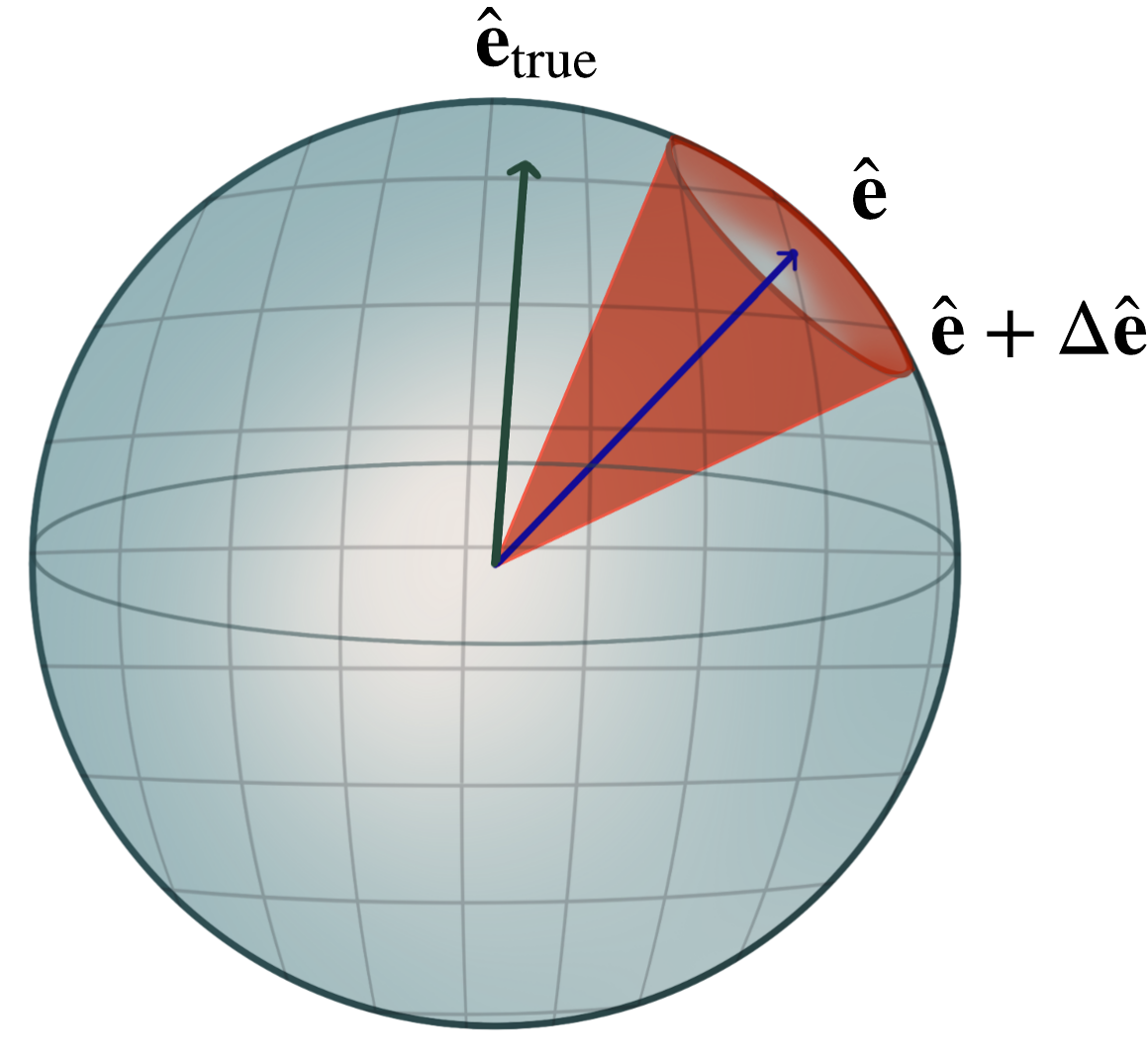} 
    \caption{Pose Sampling: Random sampling of rotation axis distributed uniformly on a spherical surface.}
    \label{fig:pose_sampling}
    \end{figure}
    
Our pose perturbation methodology is to randomly sample from the space of camera poses (i.e., rotations and translations) from the $SE(3)$ group. We bias the random pose samples to be about a reference pose vector. The pose distribution in a $\delta-$neighborhood of the reference pose ensures that the set of reference features is likely tracked in the corresponding image samples. 

Our method samples poses from all six degrees of freedom in the pose space, unlike \cite{olson2007pose}. It is important to encode all the auto and cross coupled sensitivities in the Jacobian matrix using a minimum number of perturbations. Rotations in a small $\delta-$ neighborhood of the camera axis will assist the image features to most likely be visible in the sampled images. We use axis-angle parameterization to achieve a uniform random distribution on $SO(3)$ \cite{miles1965random, shoemake1992uniform, perez2013uniform}. 
As shown in Fig. \ref{fig:pose_sampling}, the rotations are sampled from a uniform distribution of unit vectors in the unit-2 sphere, in the neighborhood of a reference axis ($\hat{\mathbf{e}}$) as $\hat{\mathbf{e}}+ \Delta \hat{\mathbf{e}}$. A non-uniform angle distribution in $[0, \theta]$ for a small $\theta$, yields uniform random rotation matrices \cite{shoemake1992uniform}. These rotation matrices are converted to CRPs ($\mathbf{q}$'s) as we use the Gibbs vector representation for orientation. 

\begin{figure}[hbt!] 
    \centering
    \includegraphics[width=0.8\textwidth]{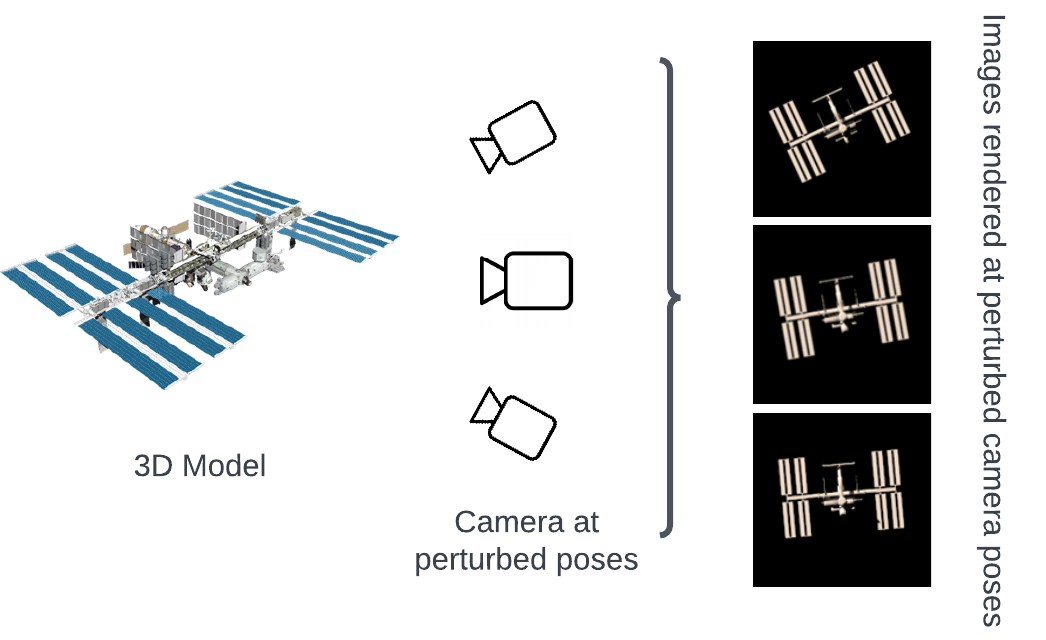} 
    \caption{Pose sampling: Renders of the target with three different camera pose perturbations.}
    \label{fig:pose_discretization}
    \end{figure}

The translational pose components are sampled from a bounded uniform random distribution along each of the axes. The distribution covers combined translations in the transverse as well as the forward motions. Our strategy is to uniformly space the samples in a bounding volume pose space such that the rendered images from adjacent samples have substantial overlap. Knowledge of perturbed camera pose is represented by $\mathbf{p} + \Delta \mathbf{p}_i$, combining sample translations for position and sample Gibbs vectors for orientation. Fig. \ref{fig:pose_discretization} illustrates the renders of a target object for three example camera poses.

Using four feature points, a vector $\mathbf{x}$ is thus defined as a set of eight feature coordinates:
\begin{equation}
    \mathbf{x} = (x_1, y_1, \hdots, x_4, y_4) 
\end{equation}

\subsubsection{Online Learning}

In this section, we discuss the learning-based estimation of the Jacobian matrix for the image feature rendering process. For illustration, we choose the specific case of an object with four key-point features, which can be tracked across all the images rendered in the perturbed pose space. However, note that the method can be applied to any number of feature points (at least three). In addition, if the tracking of at least four features is unsuccessful, the method re-initializes with conservative pose samples.

Tracking the features of the images rendered at all the perturbed poses, we numerically evaluate the local gradient of the image feature rendering process in the neighborhood of a reference pose. For all pose perturbations, we measure errors $\Delta \mathbf{x}_i$ due to feature differences between the reference image features $\mathbf{x}_{\text{ref}}$ and the perturbed image features $\mathbf{x}_i$ due to the variation in pose parameters $\Delta \mathbf{x}_i$. We write this error as follows. 
\begin{equation}
    \Delta \mathbf{x}_i = \mathbf{x}_{\text{ref}} - \mathbf{x}_i
\end{equation}

Furthermore, for all pose perturbations, we populate the difference between the reference pose $\mathbf{p}_{\text{ref}}$ and a sampled pose $\mathbf{p}_i$ by combining the differences in the rotation and translation components.
\begin{align}
     \Delta \mathbf{R}_{\text{i}} &= \mathbf{R}_{\text{ref}}^{\text{T}} \mathbf{R}_{\text{i}}
     \\
     \Delta \mathbf{t}_{\text{i}} &= \mathbf{t}_{\text{ref}}-\mathbf{t}_{\text{i}}
\end{align}

\noindent With the attitude represented using the Gibbs vector ($\mathbf{q}$), the relative pose difference can be written as $\Delta \mathbf{p}_{\text{i}} = [\Delta \mathbf{q}_{\text{i}} \quad \Delta \mathbf{t}_{\text{i}}]^T$.

Ultimately, we linearize the nonlinear transformation, denoted by $\mathbf{H}$, from the image feature space to the pose space for $N_{s}$ perturbations about the reference pose, to write 
\begin{equation}
    \underbrace{[\Delta \mathbf{x}_1, \, \hdots, \,  \Delta \mathbf{x}_{N_s}]}_{\mathbf{E}} = \frac{\partial \mathbf{H}}{\partial \mathbf{p}} \underbrace{[\Delta \mathbf{p}_1, \, \hdots, \, \Delta \mathbf{p}_{N_s} ]}_{\mathbf{B}}
\end{equation}

For a sufficiently large number of pose samples, the gradient (Jacobian) of the image feature rendering process is computed by evaluating the right inverse of $\mathbf{B}$ as
\begin{equation} \label{eq:Jac_learned}
    \mathbf{J}_f = \frac{\partial \mathbf{H}}{\partial \mathbf{p}} = \mathbf{E} (\mathbf{B}^T {(\mathbf{B} \mathbf{B}^T)}^{-1})
\end{equation}

\noindent For $N_s$ number of perturbations, the dimension of $\mathbf{E}$ is $8 \times N_s$, dimension of $\mathbf{B}$ is $6 \times N_s$ and that of $\mathbf{J}_f$ is $8 \times 6$. There are $48$ unknowns in $\mathbf{J}_f$. This model is equivalent to a least square solution of a system of $N_s$ equations with $8$ unknowns for $6$ right hand sides. The rank deficiencies in $\mathbf{B}$ can be avoided by randomly and carefully perturbing pose along all the $6$ axes.
\section{Pose Optimization Problem} \label{sec:optim}

The least squares objective function $\mathcal{L}$, for the optimization of pose in feature-based visual odometry, can be written as follows:     
\begin{equation} \label{eq:optimization_eq}
\mathcal{L}(\mathcal{S}, \mathcal{X}_p, \mathbf{p})= \min_{\mathbf{p}}  \sum_{i} ||\tilde{\mathbf{x}} - \mathbf{h}(\mathbf{s}_i,\hat{\mathbf{p}}) || ^2 
\qquad s_i \in \mathcal{S}, \,   \tilde{\mathbf{x}} \in \mathcal{X}_p
\end{equation}

\noindent where $\mathcal{S}$ is a set of RGB images, $\mathcal{X}_p$ is a set of features extracted from images in $\mathcal{S}$, $\tilde{\mathbf{x}}$ denote the features observed in the 2D image rendered at true pose. $\mathbf{h}(\mathbf{s}_i,\hat{\mathbf{p}})$ is the measurement function that captures the geometric relationship between the predicted pose $\hat{\mathbf{p}}$ and the features of the image frame $\mathbf{s}_i$ that is rendered at the predicted pose. 

Solving the objective of Eq. (\ref{eq:optimization_eq}) involves the first-order approximation of the nonlinear measurement function $\mathbf{h}(\mathbf{s}_i,\hat{\mathbf{p}})$ about an initial pose guess $\hat{\mathbf{p}}^{(\text{i})}$: 

\begin{equation} \label{eq:optimization_form}
\sum_{i} ||\tilde{\mathbf{x}} - \mathbf{h}(\mathbf{s}_i,\hat{\mathbf{p}}) || ^2 
=
\sum_{i} ||\tilde{\mathbf{x}} - \mathbf{h}(\mathbf{s}_i,\hat{\mathbf{p}}^{(\text{i})}) -  \mathbf{J}_f (\mathbf{p}-\hat{\mathbf{p}}^{(\text{i})})|| ^2 
\end{equation}

Utilizing the Jacobian in Eq. (\ref{eq:Jac_learned}) computed by online learning, we regress upon the relative pose parameters of the object with respect to the camera by minimizing the feature misalignment errors between the images synthesized at the predicted and the true poses. Levenberg-Marquardt (LM) algorithm for nonlinear least squares is used to iteratively update the pose parameters by correcting the initial pose values until convergence. The iterative procedure is represented as:
\begin{equation}{\label{eq:nonlinearOptimDR}}
    \hat{\mathbf{p}}_{i+1} = \hat{\mathbf{p}}_{i} + \left(\mathbf{J}^T\mathbf{J} + \lambda \text{diag}(\mathbf{J}^T\mathbf{J})\right)^{-1}\mathbf{J}^T \left( \tilde{\mathbf{x}} - \mathbf{h}(\mathbf{s}_i,\hat{\mathbf{p}}^{(\text{i})})\right)
\end{equation}

\noindent \noindent where $\lambda$ is the LM update parameter that adaptively modifies the reduction in error residuals.

The method minimizes the pixel differences between the reference image features (rendered at target pose) and the matching features from image rendered at the initial pose guess. It dynamically updates the pose by an amount of $\delta \mathbf{p}$ until convergence, i.e., $||\delta \mathbf{p}|| < \epsilon$ (for a small $\epsilon$). The algorithm \ref{alg:pose_est} represents the sequences of steps involved in the pose estimation procedure.

\renewcommand{\algorithmicrequire}{\textbf{Input:}}
\renewcommand{\algorithmicensure}{\textbf{Output:}}

\begin{algorithm}
\caption{Pose estimation algorithm}\label{alg:pose_est}
\begin{algorithmic}[1]
\Require Reference image $\mathbf{s}_{\text{ref}}$, initial pose guess $\hat{\mathbf{p}}_{0}$, LM update parameter $\lambda$
\Ensure Converged pose: $\hat{\mathbf{p}}_{\text{final}}$
\State Extract reference image features $\tilde{\mathbf{x}}$ from the reference image.

\While{$||\delta \mathbf{p}|| > \epsilon$}
\State Render image at initial pose guess $\hat{\mathbf{p}}^{(\text{i})}$.
\State Extract features and match them with the reference features  $\tilde{\mathbf{x}}$ as $\mathbf{h}(\mathbf{s}_i,\hat{\mathbf{p}}^{(\text{i})})$.
\State Sample pose parameters around the initial pose guess $\hat{\mathbf{p}}_{i}$. 
\State Render images at sampled pose values and match reference features in each of them. 
\State Calculate $\mathbf{J}_f$ via online learning and using Eq. (\ref{eq:Jac_learned}).
\State Calculate the least squares pose update $\hat{\mathbf{p}}_{i+1}$ using Eq. (\ref{eq:nonlinearOptimDR}).
\EndWhile
\State \Return Converged pose: $\hat{\mathbf{p}}_{\text{final}}$
\end{algorithmic}
\end{algorithm}


\section{Results} \label{sec:res}
In this section, we present simulation results to demonstrate the application of differentiable rendering to estimate pose and realistically reproduce an observed image. For illustration, we simulate two proximity navigation scenarios for pose estimation during: (a) spacecraft approach maneuver toward the International Space Station (ISS), and (b) asteroid relative navigation. The pose optimization process is run until the pose converges to be within a minimum acceptable tolerance. We define the cost of the optimization process to be the squared sum of the residuals between the image features rendered at the reference and the predicted poses. Running the iterative optimization for cost convergence is equally meaningful. Moreover, to evaluate the pose refinement process, we consider the pixel-wise distance between the reference image and the image rendered at the regressed pose value.  

\begin{figure}[hbt!]
\centering
\includegraphics[width=1\textwidth]{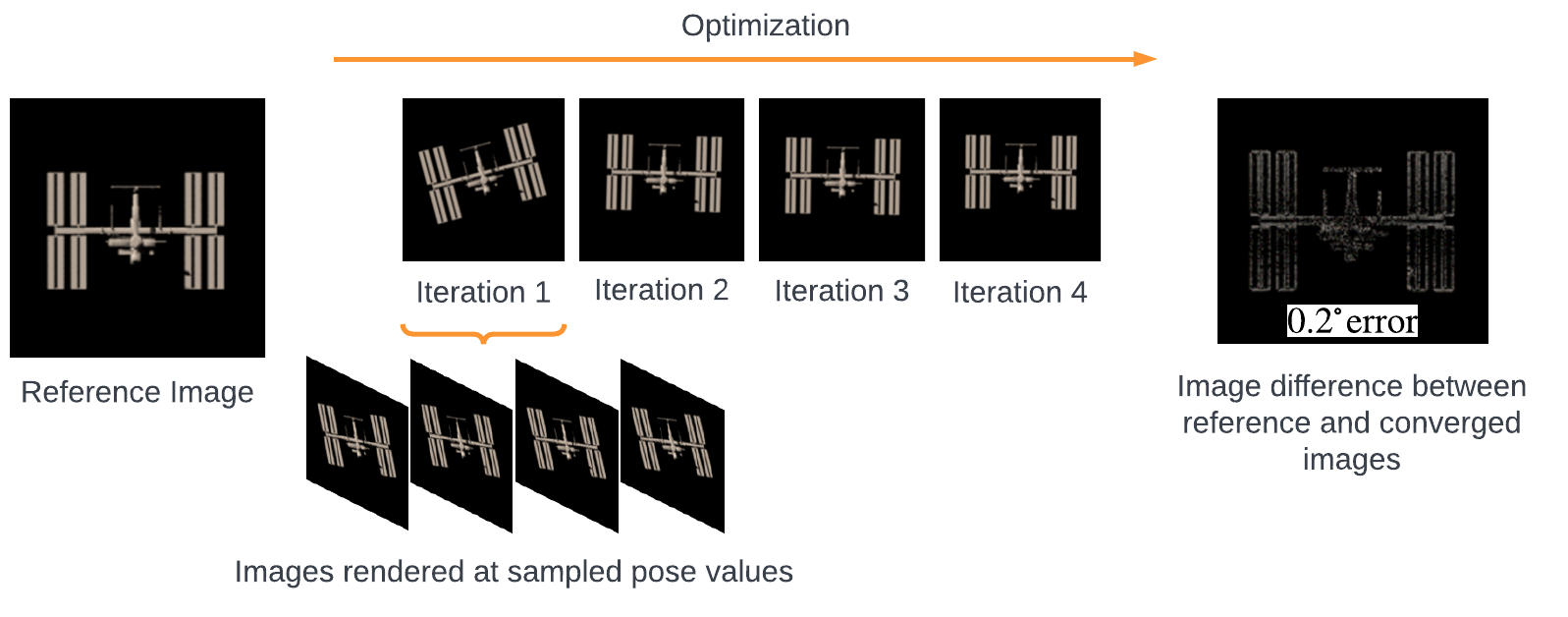}
\caption{Pose estimation and view reproduction using the differentiable rendering method. Left: A target image of the ISS synthesized at the ground-truth pose parameters. Center: minimizing the image differences with numerical gradient computed by pose perturbations. Right: Pixel differences between the image rendered at the converged pose estimate and the reference image.}
\label{fig:angle_estFig1}
\end{figure}

The ground truth is generated by forward rendering the known target model at the true pose. We estimate the pose parameters using differentiable rendering at each step to descend in the gradient direction using nonlinear optimization (Eq. (\ref{eq:nonlinearOptimDR})). Fig. \ref{fig:angle_estFig1} highlights the differentiable rendering philosophy to synthesize and estimate pose for view reproduction. The images synthesized in the first few iterations visually validate the convergence of the algorithm towards the true pose.

\subsection{Evaluation Metrics}

The consistency in the estimation of the pose can be evaluated by comparing the estimated pose with the corresponding ground truth \cite{shavit2019introduction}. Given a ground truth pose $\mathbf{p} = [\mathbf{R} \quad  \mathbf{t}]^T$ and an estimated pose $\hat{\mathbf{p}} = [\hat{\mathbf{R}} \quad \hat{\mathbf{t}}]^T$, the error in $\hat{\mathbf{p}}$ is measured by the differences in translation and orientation between $\hat{\mathbf{p}}$ and ${\mathbf{p}}$. The translation error ($||\Delta \mathbf{t}||$), measured in units of scene geometry, is evaluated by computing the absolute distance between the true and estimated translation vectors. 
\begin{equation}
    ||\Delta \mathbf{t} || = ||\mathbf{t} - \hat{\mathbf{t}}||_2
\end{equation}
The rotation error is evaluated as the angle ($\angle \Delta \mathbf{R}$) required to align the estimated and ground-truth orientations. 
\begin{equation}
    \angle \Delta \mathbf{R} = \arccos\left(\frac{\text{tr}(\mathbf{R}^{-1}\hat{\mathbf{R}}) - 1}{2} \right)
\end{equation}
\noindent Here, $\text{tr}({A})$ represents the trace of a $3 \times 3$ matrix $A$. 

\subsection{Feature Correspondence}
The differentiable rendering pipeline is illustrated in the Algorithm \ref{alg:pose_est}. The algorithm solves the feature-correspondence problem in two stages of each iteration. First, we find the correspondence between the reference image and the image rendered at the initial pose guess. Next, we find a feature correspondence between the reference image and the multiple images rendered at the sampled pose values. 

To realize feature correspondences, robust geometric feature detectors \cite{bay2006surf, shi1994good}, consistent with the type of image regions, are deployed to extract feature descriptors. Using the descriptors, the features of both images are matched. Blocking as many mismatched pairs as possible using outlier rejection, the best few feature matches between the images are selected as sparse measurements. 

The feature matching process is computationally inefficient but unavoidable when the two images are rendered from cameras that are far apart. However, the images rendered at the sampled poses are from camera views that are very close. Here, the Kanade-Lucas-Tomasi (KLT) feature-based tracking algorithm \cite{lucas1981iterative} allows for efficient and real-time feature correspondence between the reference image and the large number of views rendered at the sampled pose values.


\subsection{Experiments}

Two experiments are described to validate the performance of the proposed method to estimate the 6-DoF pose. In both the experiments, we start the optimization using an informed guess about the pose parameters. The guess has to be meaningful enough to find a minimum number of feature matches between the reference and the image rendered at the guessed pose. 
\begin{enumerate}
    \item In the first experiment, starting from a random initial pose, we reproduce a reference image of the International Space Station (ISS) captured at a reference pose. In the process, we optimize the initial pose and converge toward the reference pose at which the reference image is rendered.
        We quantitatively evaluated the pose convergence in Fig. \ref{fig:iss_3_stats} and qualitatively illustrated the image convergence in Fig. \ref{fig:iss_3_iterations}. Translation and rotation errors after 10 iterations are observed to be $0.12$ m and $0.116 ^\circ$, respectively. 
\newline
Fig. \ref{fig:iss_3_pxDiff} shows the absolute image difference between the reference ISS image and the image synthesized at the converged pose. Pixel differences indicate not only shape mismatches, but also illumination differences. In particular, the bright left half in Fig. \ref{fig:iss_3_pxDiff} hints that the position of the illuminating source to be towards the left of the target object (ISS). 
        
\begin{figure}[H]
     \centering
     \subfloat[][Rotation errors (degrees)]{\includegraphics[width=.33\textwidth]{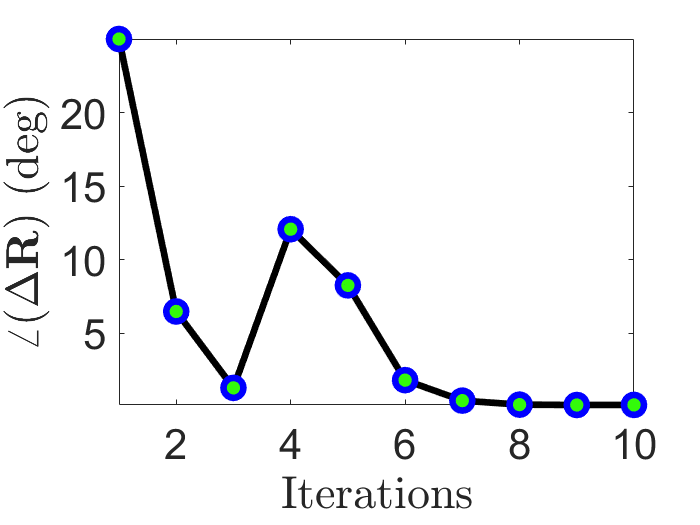}\label{fig:case3_attitude}}
     \subfloat[][Translation errors (meters)]{\includegraphics[width=.33\textwidth]{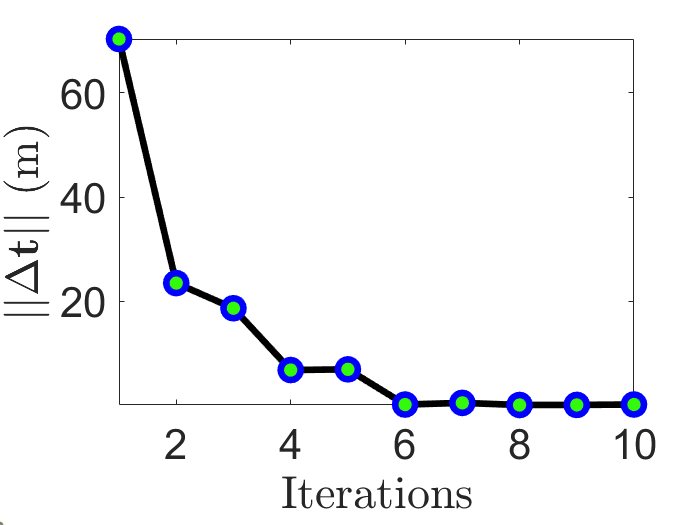}\label{fig:case3_transl}} 
     \subfloat[][Logarithmic cost]{\includegraphics[width=.33\textwidth]{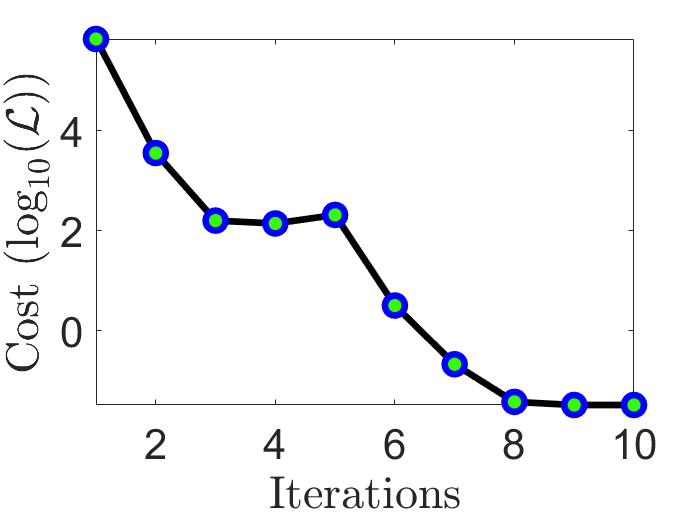}\label{fig:case3_cost}}
     \caption{Experiment 1 ("ISS"): Error metrics evaluated at each iteration of the pose optimization.}
     \label{fig:iss_3_stats}
\end{figure}

\begin{figure}[H]
     \centering
     \subfloat[][Iteration 1]{\includegraphics[width=.25\textwidth]{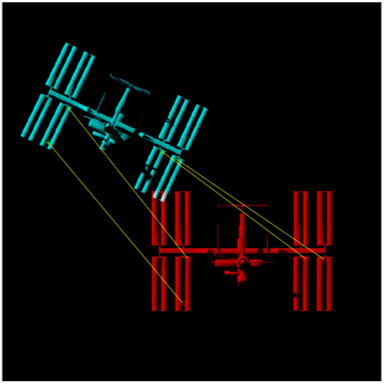}\label{fig:iss_3_iter1}}
     \subfloat[][Iteration 2]{\includegraphics[width=.25\textwidth]{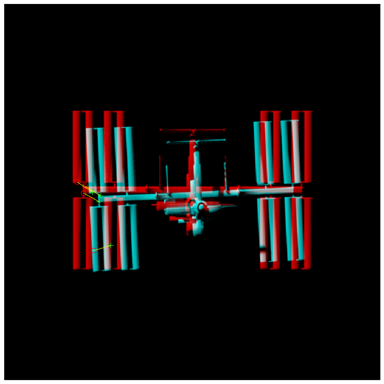}\label{fig:iss_3_iter2}} 
     \subfloat[][Iteration 10]{\includegraphics[width=.25\textwidth]{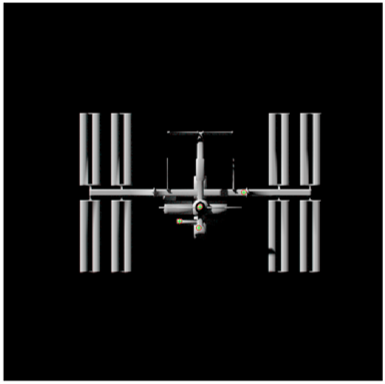}\label{fig:iss_3_iter10}}
     \subfloat[][Image difference]{\includegraphics[width=.25\textwidth]{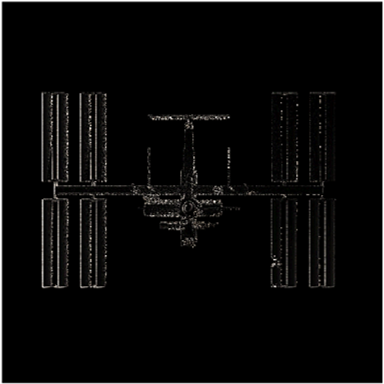}\label{fig:iss_3_pxDiff}}
     \caption{Experiment 1 ("ISS"): Optimization iterations in matching an image rendered at a random initial pose (blue) with the reference image (red). The fourth image represents pixel wise difference between the converged and the reference images.}
     \label{fig:iss_3_iterations}
\end{figure}

\item The second experiment involves navigation relative to the terrain of "Rheasilvia" \footnote{The Rheasilvia basin: \url{https://nasa3d.arc.nasa.gov/detail/vesta-rheasilvia}}. Starting from an initial pose guess, we match the reference image and also converge towards the target pose. 
    We quantitatively evaluated the pose convergence in Fig. \ref{fig:rh_3_stats} and qualitatively illustrated the image convergence in Fig. \ref{fig:rh_3_iterations}. The translation and rotation errors obtained are observed to be $2.38$ m and $2.11 ^\circ$, respectively. 
    \newline
    Fig. \ref{fig:rh_3_pxDiff} shows the absolute image difference between the reference image and the image rendered at the converged pose. Noticeably, the pixel differences indicate not only the shape differences but also the illumination differences. 

\begin{figure}[H]
     \centering
     \subfloat[][Rotation errors (degrees)]{\includegraphics[width=.33\textwidth]{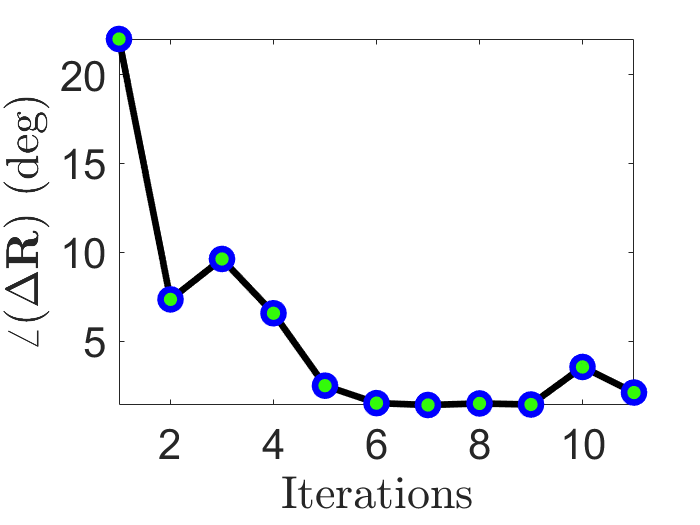}\label{fig:case4_attitude}}
     \subfloat[][Translation errors (meters)]{\includegraphics[width=.33\textwidth]{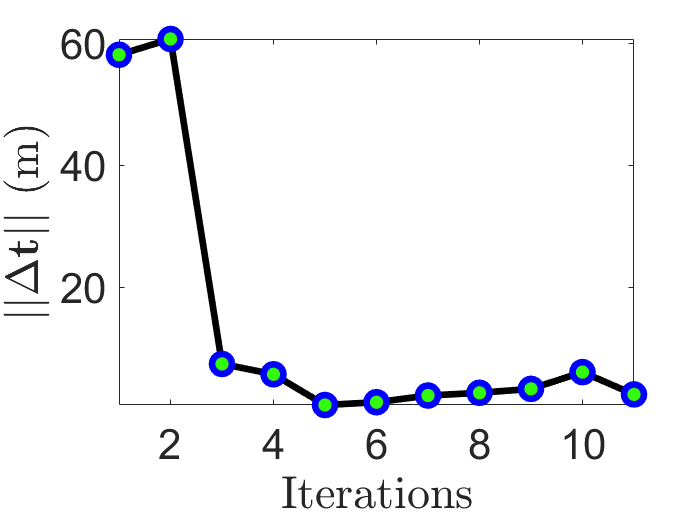}\label{fig:case4_transl}} 
     \subfloat[][Logarithmic cost]{\includegraphics[width=.33\textwidth]{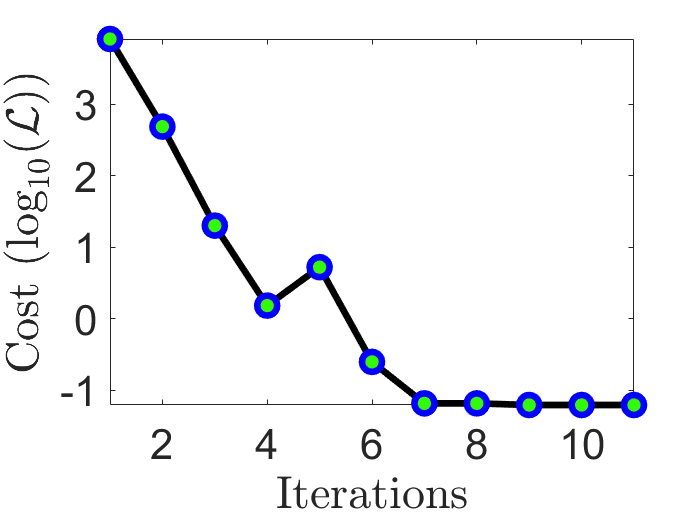}\label{fig:case4_cost}}
     \caption{Experiment 2 ("Rheasilvia"): Error metrics evaluated at each iteration of the pose optimization.}
     \label{fig:rh_3_stats}
\end{figure}

\begin{figure}[H]
     \centering
     \subfloat[][Iteration 1]{\includegraphics[width=.25\textwidth]{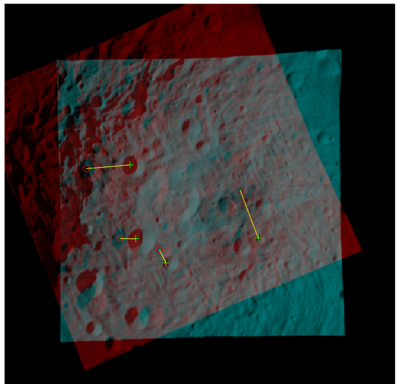}\label{fig:rh_3_iter1}}
     \subfloat[][Iteration 2]{\includegraphics[width=.25\textwidth]{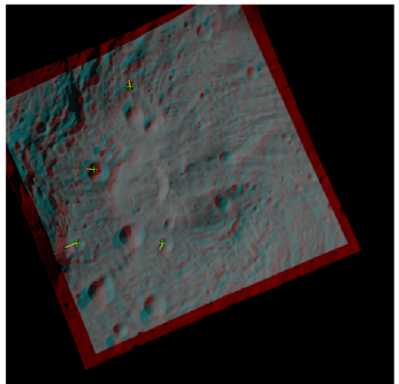}\label{fig:rh_3_iter2}} 
     \subfloat[][Iteration 11]{\includegraphics[width=.25\textwidth]{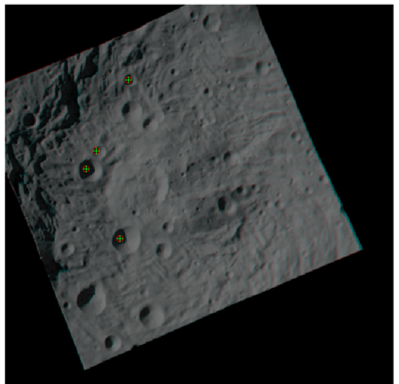}\label{fig:rh_3_iter10}}
     \subfloat[][Image difference]{\includegraphics[width=.25\textwidth]{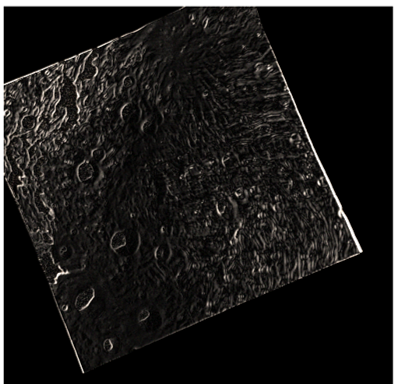}\label{fig:rh_3_pxDiff}}
     \caption{Experiment 2 ("Rheasilvia"): Optimization iterations in matching an image rendered at a random initial pose (blue) with the reference image (red). The fourth image represents pixel wise difference between the converged and the reference images.}
     \label{fig:rh_3_iterations}
\end{figure}

\end{enumerate}

\section{Conclusion}

In this paper, we have proposed a pipeline to estimate the 6-DoF pose of a target from a single image. Starting from an initial pose guess, we rendered multiple images by perturbing the guess, and thereby learned the local gradient of the image formation model in a derivative-free approach. By computing only the 2D feature differences, we obtained a least-squares estimate for the 6-DoF pose using differential corrections to the initial pose guess. To show the validity of the proposed algorithm, we presented two proximity operation experiments. 

Learning the local gradient of the image formation is feasible with the knowledge of the target's 3D model. The availability of the model enables the rendering of images at different pose perturbations to facilitate feature-rich datasets for online learning. However, if a model is not readily available for an uncooperative target, a dense point cloud can be constructed from a LiDAR scanner or synthesized from multiview 3D reconstruction. The point cloud facilitates solving the 2D feature to 3D coordinate correspondence for pose estimation (refer to our previous work \cite{eapen2022narpa}).

The proposed pipeline is sensitive to the establishment of a minimum number of 2D feature correspondences. Future work will be aimed at researching feature matching algorithms relevant to a target object's texture, outlier rejection schemes, and guarantees for local minima. Image rendering is the computationally expensive step in the algorithm. For real-time implementation, high-speed computing architectures are to be explored. 

Furthermore, reasoning about the optimal pose sampling methods that accurately capture the gradient of the image formation model, remains an open research problem.

\bibliography{references}

\end{document}